%% file: acl_latex.tex
\pdfoutput=1
\documentclass[11pt]{article}

\usepackage[]{acl}

\usepackage{times}
\usepackage{latexsym}
\usepackage{tcolorbox}
\usepackage{tikz}
\usetikzlibrary{positioning, shapes, fit}
\usepackage{fontawesome5} 
\usepackage{pifont}
\usepackage{multirow} 
\usepackage{lipsum}
\usepackage{hyperref}
\usepackage{booktabs} 

\usepackage[T1]{fontenc}
\usepackage[utf8]{inputenc}

\usepackage{microtype}

\usepackage{inconsolata}

\usepackage{graphicx}
\usepackage{soul}

\usepackage{subcaption}
\usepackage{xspace}
\usepackage{todonotes}
\newcommand*\iftodonotes{\if@todonotes@disabled\expandafter\@secondoftwo\else\expandafter\@firstoftwo\fi}  % defines \iftodonotes{<true>}{<false>}, thanks to \makeatother

\usepackage{color-edits}
\addauthor{ZT}{olive}
\addauthor{MSS}{blue}
\addauthor{CdK}{yellow}
\addauthor{PM}{green}

\title{IYKYK: Using language models to decode extremist cryptolects}

\author{Christine de Kock\thanks{\textbf{Correspondence:} \href{mailto:christine.dekock@unimelb.edu.au}{christine.dekock@unimelb.edu.au}}\\
  University of Melbourne \\
  \And
  Arij Riabi\\
  Inria, Paris, France\\
  \AND
  Zeerak Talat\\
  University of Edinburgh\\
  \And
  Michael Sejr Schlichtkrull \\
  Queen Mary University of London\\
  \AND
  Pranava Madhyastha\\
  City St George's, University of London\\
\And
  Eduard Hovy\\
  University of Melbourne\\
  }

\begin{document}
\maketitle
\begin{abstract}
Extremist groups develop complex in-group language, also referred to as \textit{cryptolects}, to exclude or mislead outsiders. We investigate the ability of current language technologies to detect and interpret the cryptolects of two online extremist platforms. Evaluating eight models across six tasks, our results indicate that general purpose LLMs cannot consistently detect or decode extremist language. However, performance can be significantly improved by domain adaptation and specialised prompting techniques. These results provide important insights to inform the development and deployment of automated moderation technologies. We further develop and release novel labelled and unlabelled datasets, including 19.4M posts from extremist platforms and lexicons validated by human experts.
\end{abstract}

\section{Introduction}
The proliferation of online extremist ideologies, often veiled in complex cryptolects, poses a severe societal threat and a unique challenge for Natural Language Processing (NLP). Although online spaces have been used to facilitate the activities of extremist groups since shortly after the creation of the world wide web \citep{strf,conway2006terrorism}, concerns over the recent rise of online extremist activity have been expressed by researchers \citep{baele2021incel}, the media \citep{rich2023}, NGOs \citep{latimore2023incels}, and governments \citep{pr_downingstr} alike. 

\begin{figure}[t]
    \input{examples}
    \caption{Three posts from the IYKYK  dataset (in gray) with translations by \texttt{Llama-3.3} (in green). The posts originate from Incels (\#1 \& \#2) and Stormfront (\#3).}
    \label{fig:posts_examples}
\end{figure}\looseness=-1

While modelling hate speech and toxic content are well-studied tasks within NLP, the related problem of modelling extremist language has received comparatively little attention. The latter has distinctive characteristics that make it challenging: the vernaculars of different ideologies are very specialised, often subtle, and highly dynamic \citep{weimann2020digital}. \citet{bogetic2023race} states that ``the neologisms being produced by Incels have come to form a true cryptolect, developing at a rate that almost escapes linguistic description''. In addition to neologisms, extremist groups also create ``dogwhistles'' by imbuing existing words with a coded meaning unknown to the general population \citep{mendelsohn2023dogwhistles}. Furthermore, all extremist language is not necessarily hateful: specialised language may be used to signal in-group status while discussing benign topics \cite{mcculloch2019because}, meaning that sentence context is not always helpful. \looseness=-1

The result is that, although these groups often interact on public discussion forums, their language can be incomprehensible to outsiders. Furthermore, it is not sufficient to be able to detect extremist language: being able to decode and interpret it is essential. Fig.~\ref{fig:posts_examples} illustrates this problem, showing three posts with translations by \texttt{Llama 3.3} \citep{grattafiori2024llama}. In the first post, we note the phrase \textit{going ER}, which refers to committing a mass murder targeting women\footnote{\href{https://en.wiktionary.org/wiki/go_ER}{ER} refers to Elliot Rodger, a mass murderer and icon within the incel community.}, and the term \textit{roping}, which refers to suicide. Both terms are inaccurately decoded by the model, illustrating the importance of understanding the limitations of these models. 

Complicating matters, toxic language is typically removed during pretraining of modern NLP models~\citep{anil2023palm2technicalreport, longpre-etal-2024-pretrainers} to prevent the generation of toxic content. Removing such data can even improve performance on non-toxic domains~\citep{soldaini-etal-2024-dolma}. This prevents models from generating toxic content, but could limit their ability to model extremist cryptolects.

The central hypothesis of this work is that generalised language technologies require specialised techniques when applied to extremist cryptolects. We evaluate this hypothesis in two ways. Firstly, we investigate the ability of instruction-tuned models to recognise and interpret extremist language in a zero-shot setting. Secondly, we evaluate the impact of domain adaptation for supervised classification with fine-tuned encoder models. To enable these experiments, we create a dataset\footnote{All models and datasets produced in this work will be made available to verified researchers upon request.} of 19.4M posts from two online extremist forums, Stormfront and Incels, that are dedicated to white supremacist and misogynistic ideologies, respectively. We further create test sets for three tasks using this data, employing human experts for annotation. 

These test sets are used in our zero-shot evaluation to compare seven instruction-tuned large language models (LLMs). For the task of identifying in-group language, the top models achieve an F1 of 64.4\% and 80.0\% for Stormfront and Incels, respectively. For both platforms, adding context (in the form of post examples) increases recall while decreasing precision, while adding more extensive instructions yields an F1 increase of 10\%. For the task of generating definitions, adding post examples improves performance for both platforms, with 90.3\% of definitions being correct for Incels when including 10 posts, while only 44.7\% are correctly generated when no context is provided. All models perform better at identifying and defining in-group language for Incels compared to Stormfront, which is impacted by the former's tendency to reuse common morphemes when creating new words. 

For detecting radical content, we find that the domain adaptation of encoder LLMs consistently improves performance across two tasks, achieving state-of-the-art results in English. In particular, fine-tuning \texttt{xlmt} on our dataset significantly boosts macro-F1 scores (+3.6\% and +5.2\%), especially when combined with auxiliary task such as ideology prediction. Interestingly, the results indicate that performance is improved even when the domain adaptation uses data from an unrelated extremist ideology. Finally, we find that domain adaptation on our data also improves the performance of classifiers for hate speech detection. 

\section{Background}
\subsection{Hateful in-group language}
Our work is grounded in two recent works in NLP that investigate the use of specialised language to convey a specific (often hateful) message to an in-group. We detail their approaches here.

\citet{mendelsohn2023dogwhistles} investigate the topic of dogwhistles using a newly-created glossary of 300 antisemetic dogwhistles. They evaluate whether an LLM can correctly provide dogwhistles’ covert meanings, finding that providing additional cues in the prompt (such as the definition of the concept of a dogwhistle) improves performance. Secondly, they investigate the ability to surface dogwhistles, prompting the model to generate examples based on the definition and comparing the output with their glossary. Finally, they perform a small-scale manual analysis of in-context dogwhistle recognition over 5 samples, prompting the model to identify dogwhistles given a real-world example usage. They note that a larger-scale, more rigorous analysis is warranted for the latter task. 

Secondly, \citet{de2024listn} introduces an approach for inducing lexicons of extremist in-group language based on social and temporal characteristics of online conversations. Manual validation by experts is used to create a test set. Given a word and usage examples, experts are tasked with assigning a binary label representing whether the word constitutes in-group language of the group in question. In this work, we evaluate the ability of LLMs to replicate this task, which is taxing on human moderators. We use their lexicon induction and validation approaches as a basis for finding candidate words. While their approach relied primarily on social and statistical cues, we focus on language models in our evaluation.   

To our knowledge, our study is the first to perform an in-depth exploration of the ability of LLMs to identify and decode extremist in-group language.
 
\subsection{Detecting radical content}
Online extremism is frequently fuelled by exposure to radical content \citep{valentini2021digital,shaw2023social}. The availability of datasets for detecting radical content remains limited, both in terms of the range of ideologies and the representation of linguistic diversity \citep{10.1155/2023/4563145,Berjawi2023}. 

One major challenge stems from the difficulty of clearly defining what constitutes extremism or radicalization, which is a long-debated topic within counterterrorism research \citep{della2012guest}. These ambiguities make annotation particularly challenging and can introduce biases in how datasets are constructed \citep{Gaikwad2021OnlineED}. 
In this work, we use the term \textit{extremism} to refer to ideologies that promote violence against an out-group. Following \citet{riabi2025beyond}, \textit{radical content} is understood to refer to texts and other media expressing expressing a radical perspective in opposition to a political, social, or religious system.\looseness=-1

The recently introduced Counter dataset \citep{riabi2025beyond} provides annotations of radical content in English, French, and Arabic, representing primarily Jihadist and alt-right ideologies. In \citet{riabi2025beyond}, pretrained encoder models are fine-tuned to perform classification on a range of tasks, such as inferring the radicalization level and calls for action. However, the concepts underlying radicalization remain sensitive to the language used, the specific ideological context, and the particular forums in which the content is produced. In this work, we use two tasks in the Counter dataset to investigate the effect of language variation on pretrained encoder models. 

\section{The IYKYK dataset}\label{sec:data}
\begin{table}[htb]
    \centering
    \footnotesize
    \begin{tabular}{|l|l|l|}
    \hline
     & \textbf{Incels} & \textbf{Stormfront}\\
     \hline
       Start date & 2017-11-08 & 2001-09-11 \\
       \# posters in 2024 & 3827 & 1784 \\
       \# months of data & 85 & 279 \\
       \# posts  & 10.3 MM & 9.3 MM\\
       \# posters  & 16 654 & 97 430\\
       \# posts / posters &25 & 3\\
       \# posts / thread &14& 5\\
       \# char / post &44& 185\\
       \hline
    \end{tabular}
    \caption{Characteristics of the dataset. Where relevant, medians are used to calculate averages.}
    \label{tab:data_char}
\end{table}
To enable our zero-shot and domain adaptation experiments, we construct a new dataset of posts from Incels and Stormfront. The dataset consists of all publicly available posts on both platforms up to November 2024, totalling 19.4 million posts. For each post, we include thread ID, platform subcategory, user ID, and posting date as meta data. Our dataset only includes users who have posted on the platforms and therefore does not reflect the total number of users or visitors of each platform\footnote{The number of active unregistered visitors, as reported on the respective homepages, is often up to 100 times more than the active registered users, with Stormfront reporting a maximum of 282\,293 online users on 30 November 2022.}.\looseness=-1 

We note that there are differences between the two communities in our dataset (see Table \ref{tab:data_char}). Specifically, users on Incels tend to be more active, with each user posting an average of 25 posts, compared to 3 on Stormfront. Posts on Stormfront, however, are on average 4.2 times longer than on Incels, while Incels features longer threads (2.8x). Using the \texttt{langid} package, we find that the posts are predominantly in English (95\% for Stormfront and 87\% for Incels) with minor French, Spanish, and German components in both; however, we expect that there is some noise in this result due to the non-standard nature of the language. \looseness=-1 

\section{Experimental setting}\label{sec:methods}
We examine two different settings to evaluate the applicability of current dominant language technologies to extremism-related tasks. In the first part, we investigate the utility of zero-shot evaluation of seven instruction-tuned models for \emph{classifying} in-group language, \emph{retrieving} instances of in-group language, and \emph{decoding} (i.e., defining) in-group terminology. In the second part, we perform supervised classification, using \texttt{xmlt} to evaluate the \emph{domain adaptation} capabilities of large pre-trained language models for detecting radical content.

\subsection{Zero-shot evaluation}\label{sec:tasks}
\begin{figure}[]
    \centering
    \includegraphics[width=0.9\linewidth]{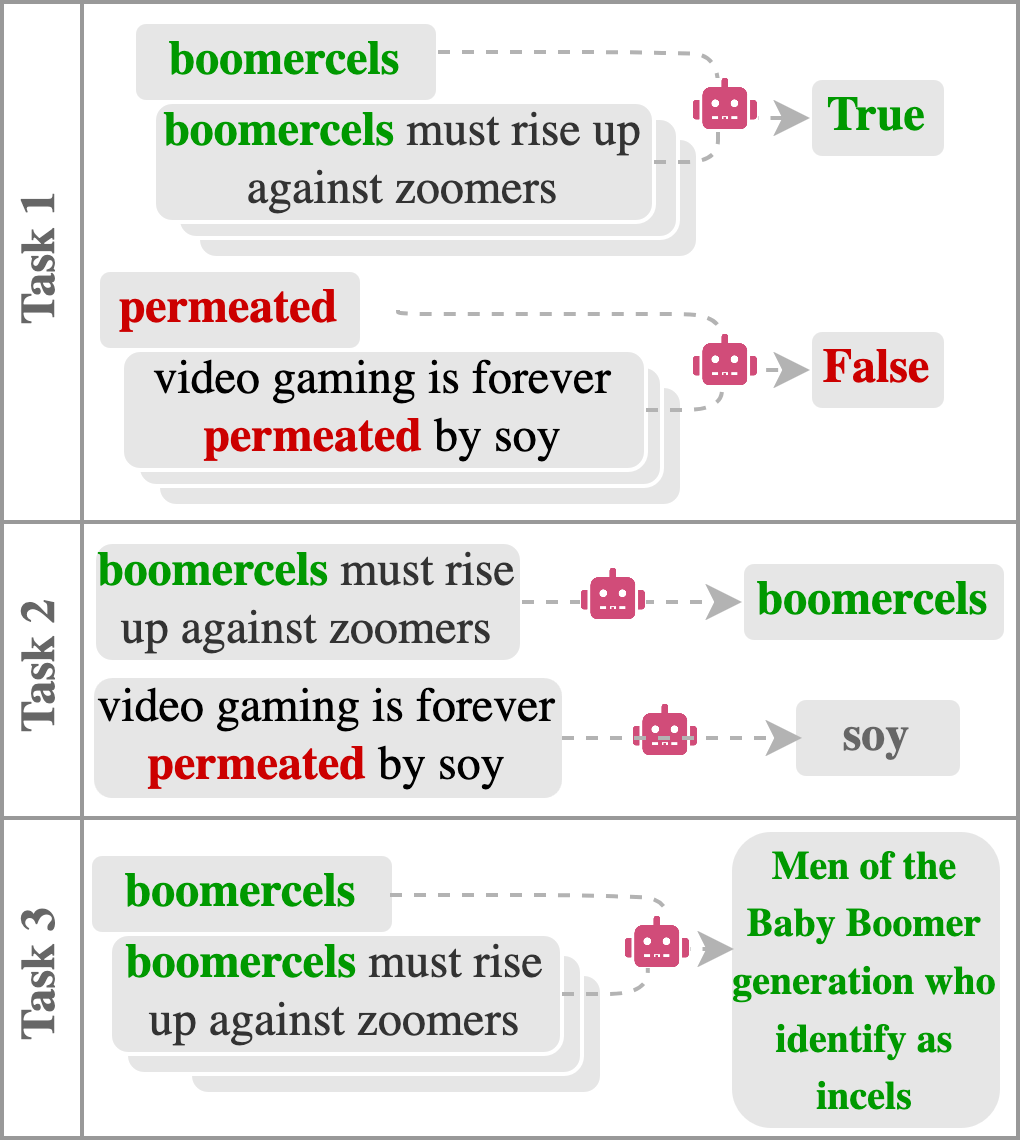}
    \caption{Tasks used for the zero-shot LLM evaluation.}
    \label{fig:tasks}
\end{figure}

Fig.~\ref{fig:tasks} illustrates the tasks we use to evaluate the instruction-tuned models. Seven models are used: \texttt{Gemma 2} (27B and 9B parameters, \citealp{team2024gemma}), \texttt{Llama 3.1} and \texttt{3.3} (8B  and 72B, \citealp{grattafiori2024llama}), \texttt{Mixtral 8x} (7B, \citealp{jiang2024mixtral}), and \texttt{Qwen 2.5} (7B and 72B, \citealp{bai2023qwen}). For each task, we experiment with four context framing strategies in the prompt, as detailed below. In Sec.~\ref{sec:lex_ind} , we describe the pipeline for generating test cases and the human experts employed to assign gold standard labels. Full prompts and implementation details are provided in App.~\ref{app:prompts}. 

\paragraph{Task 1: Classification}
Given a target word, each model is tasked with evaluating whether it constitutes valid in-group language for a particular community. Here, we closely emulate the human validation setup as described in \citet{de2024listn}. For each platform, 3050 test cases, including positive and negative examples, are evaluated. We report the F1 scores.

\paragraph{Task 2: Retrieval} Task 2 requires the model to identify all in-group language within a given post, contextualizing a specific target word (e.g., from the relevant platform). This is more demanding than Task 1 due to the higher ratio of standard to specialized words in typical posts. However, evaluation strictly focuses on the target word: we evaluate if known positive targets are included in the model's output and confirmed negative targets are excluded. For instance (see Fig.~\ref{fig:tasks}, Task 2), if \textit{permeated} (a negative sample) is the target, and the model flags \textit{soy} as in-group, our evaluation only checks that \textit{permeated} is not flagged. While similar to the retrieval task of \citet{mendelsohn2023dogwhistles}, our approach enables larger-scale investigation and incorporates confirmed negatives, a need they identified. We again use F1 score as metric.

\paragraph{Task 3: Decoding} Given a target word, the model is tasked with providing its definition. For this task, we only evaluate confirmed positive instances of in-group language. Definitions are generated by the instruction-tuned \texttt{Llama-3.3} (70B) model. For each platform and context framing, 300 test cases are evaluated, totalling 2400 definitions. Human experts assess the validity of the generated definitions using the labels \textit{correct}, \textit{partially correct}, and \textit{incorrect}. The full instructions are provided in App.~\ref{app:annotation_instructions}. Test cases are drawn from the positive samples in Tasks 1 and 2 as well as prior lexica by \citet{de2024investigating}, aiming to approximately match frequency distributions of the included words between the platforms. We report as metric the validity distribution for each setting. 

\paragraph{Context framing}
We explore four levels of contextual input in the prompts. The \texttt{definition} framing parallels the setting of \citet{mendelsohn2023dogwhistles}, providing only the definition for in-group language and the target word. This probes whether the model training data included sufficient information for the model to perform the task without further context. The \texttt{instructions} framing provides the full instructions as used for the human annotation (detailed in Sec.~\ref{sec:lex_ind}) which also includes the nature of the research (focusing on extremism), the platform name and ideology, and examples of positive and negative types. The \texttt{10-examples} framing provides the full instructions and ten posts containing the target word on the relevant platform. Since there is a disparity in the post lengths between the two platforms, we select only posts with 10-300 characters. Finally, the \texttt{1-example} framing provides full instructions and one example post only. For Task 2, providing examples is infeasible, since a target word is to be retrieved from a given post. As such, we evaluate only performance with and without the extended instructions. 

\subsection{Supervised classification}
To investigate the effect of domain adaptation on radical content detection, we consider two supervised classification tasks from the Counter dataset. Firstly, {\tt Call for Action} assesses whether the content indicates a need for intervention, using a five-point scale ranging from \textit{no action needed} to \textit{urgent action}. Secondly, {\tt Radicalization Level} which captures the degree of radicalization expressed in the content. It includes six levels: \textit{Neutral}, \textit{Expression of Radical Views}, \textit{Use of Radical Propaganda}, \textit{Affiliation with Radical Groups}, \textit{Dehumanization of the Other}, and \textit{Call for Action against others}. 
We experiment with the other post-level annotations, e.g. named entities and ideological labels, as auxiliary tasks. We adopt the same 70:20:10 data split as \citet{riabi2025beyond}. Each language subset, French and English, includes 2650 annotated examples. Matching the original work, we report the macro-F1 score.

\paragraph{Models}
We experiment with {\texttt{xlmt}\xspace} \citep{barbieri-etal-2022-xlm}, an {\texttt{xlmr}\xspace} \citep{conneau-etal-2020-unsupervised} model which has been fine-tuned for the Masked Language Modelling (MLM) task on multilingual social media data. This model was used in the work of \citet{riabi2025beyond} to establish a benchmark for the {\tt Call for Action} task. As an encoder-only model, its architecture is more appropriate for the supervised classification task compared to the instruction-tuned decoder-only models used in Sec.~\ref{sec:tasks} \citep{DBLP:journals/corr/abs-2308-10092,edwards-camacho-collados-2024-language}. We use the MaChAmp v0.2 toolkit \cite{van-der-goot-etal-2021-massive}, which supports a multi-task setting with a shared encoder and separate classification heads for each task. These heads are jointly fine-tuned during training, and the encoder parameters are updated based on the combined gradients from all tasks. We report hyperparameters in App.~\ref{app:radical_content}.

\paragraph{Domain adaptation}
We aim to adapt the base {\texttt{xlmt}\xspace} encoder to the language characteristics specific to these online platforms and assess how this adaptation impacts downstream tasks within the Counter dataset. We use the IYKYK dataset to continue training the encoder model using the MLM objective, before fine-tuning it for classification as described above. 
After filtering out one-word or empty posts, we include approximately 9M posts from each forum. We experiment with three settings: fine-tuning on each forum separately and fine-tuning on the combined posts from both forums. 
In each case, we split the data into 90\% for training and 10\% for validation. Hyperparameters are reported in App.~\ref{app:radical_content}.\looseness=-1

\section{Lexicon induction pipeline}\label{sec:lex_ind}
The tasks in Sec.~\ref{sec:tasks} require confirmed positive and negative examples of in-group language for each community. While existing lexicons provide a basis for positive examples, these lexicons tend to rely on out-of-date data or are limited in size. To construct a gold standard reference of in-group language for each community, we follow the approach of \citet{de2024listn}, which entails combination of statistical methods and human expert labelling.

\paragraph{Candidate generation} 
In-group language constitutes a small fraction of the general vocabulary of a group; as such, automated methods have been developed to extract it. We use the \texttt{LISTN-C} method \citep{de2024listn}, which allows us to identify words that, based on their usage patterns, have a higher likelihood of being particularly relevant to the group. We extract the top 3050 words per community, using a lower threshold to ensure that it will include both positive and negative test cases. The system is trained separately for each community, using annual data snapshots from 2018 to 2024 from the IYKYK dataset. Further implementation details are provided in App.~\ref{app:annotation_instructions}.\looseness=-1 

\paragraph{Expert annotation}
Given a set of candidate words for each community and 10 usage examples\footnote{The examples coincide with those used in Sec.~\ref{sec:tasks}.} for each word, human experts are employed to assess whether the word constitutes valid in-group language for the community in question. \citet{drake1980social} states that a primary purpose of in-group language is to signal cohesion and solidarity with a group. As such, our expert labelling is based on the question: \textit{If you saw someone use this word in a neutral setting, would it increase your expectation that they belong to the relevant community?} This would be strongly indicative, but is not a strict requirement for inclusion. Our full annotation instructions, including examples of positives and negatives, are provided in App.~\ref{app:annotation_instructions}. The annotators are doctoral scholars in sociology who specialise in the online rhetoric of these groups. They are employed as research assistants in this project through our institution, and receive a salary of \$37/hour. A pilot annotation round is performed by one author and the annotators over 50 samples each, yielding a Cohen's Kappa of 0.76 and 0.79, respectively. 

\paragraph{Gold standard in-group language} 
The full candidate sets are annotated by one expert annotator each, resulting in 1401 positive test cases for Incels and 399 for Stormfront. The remainder of the 3050 samples for each community are confirmed negatives. 
The discrepancy in the number of positive instances per community is substantial, but not unexpected: \citet{bogetic2023race} states that the incels are more linguistically innovative compared to the alt-right. 
Furthermore, we note that the Incels vocabulary is inflated due to a set of morphemes that are repeatedly used to derive new, somewhat obscure compound words, some of which are shown in Table \ref{tab:morphemes}. 
Of the 1401 positive terms for Incels, 570 include the substrings \textit{pill}, \textit{mog}, \textit{maxx}, \textit{cel}, \textit{chad}, \textit{cuck}, and \textit{oid}. 
This has been observed in prior work: \citet{yoder2023identity} detect over 1500 unique words that include the suffix\textit{ -cel}, used to refer to an incel with some characteristic (e.g., \textit{tallcel} might refer to an incel who is tall). 

 \begin{table}[t]
    \centering
    \footnotesize
    \begin{tabular}{p{0.05\linewidth}p{0.88\linewidth}}
       \textbf{\textit{chad}}: & \textit{chadstralian, chadlike, chadland, chadsplaining, chadsexuality, subchads, chadspawn, chadstein, chadless, chadface, chadfisher, chadder, chadstruck, chaddier, chadworship, terrachad, uberchad ...}\\
        \textbf{\textit{(f)oid}}: & \textit{australoid, schizoids, roastoid, rightoids, noodlefoid, angloids, caucasoids, normoid, mayofoids, westoids, leftoid, swarthoid, foidish, insectoid, brownoid ...} \\
     \end{tabular}
    \caption{Variants of popular subwords on Incels.}
    \label{tab:morphemes}
\end{table}

\section{Results for zero-shot evaluation}\label{sec:tasks_results}

\subsection{Task 1: Classification}

\begin{figure*}[t]
    \centering
    \includegraphics[width=\linewidth]{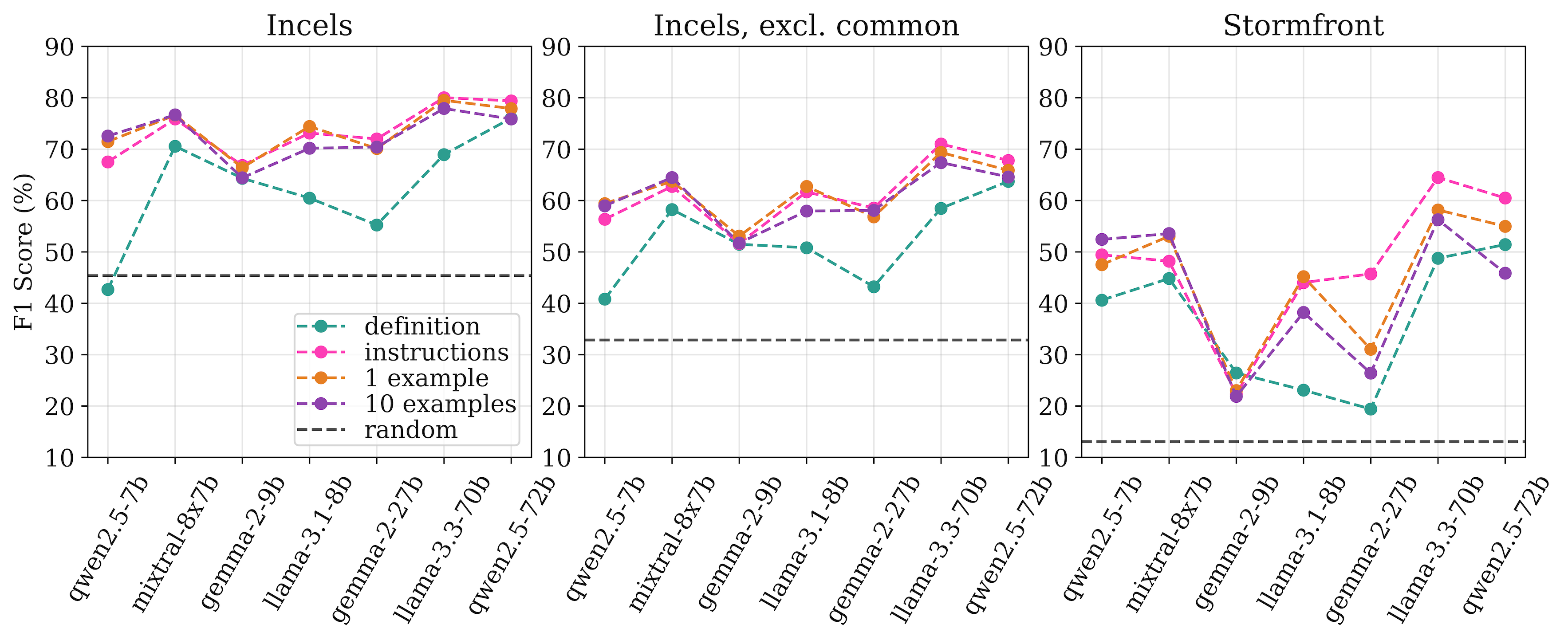}
    \caption{Results for Task 1 (Classification).}
    \label{fig:results_task1}
\end{figure*}
For both platforms, the highest F1 score (64.4\% for Stormfront and 80.0\% for Incels) is achieved by \texttt{Llama-3.3} (70B). 
Providing full \texttt{instructions} leads to increases in F1 of 9.5\% and 10.0\%, respectively, over the \texttt{definition}-only framing. Adding additional context (i.e., example posts) does not consistently lead to further increases in F1, but it does result in higher recall (shown in Fig.~\ref{fig:task1_recall} in App.~\ref{app:results}) for most models and both platforms, while precision (Fig.~\ref{fig:task1_precision} in App.~\ref{app:results}) suffers. 
In other words: including the post examples in the prompt increases the probability that the model will assign a positive label. This may be caused by the fact that the prompt indicates that the posts were sourced from the platform in question, creating a link to the group. As expected, the larger model variants tend to perform better than their smaller counterparts.

Matching model and context settings, the F1 scores are higher for Incels (left) than for Stormfront (right). This is likely impacted by the larger imbalance in the test set, also reflected in the random baseline. The commonly reused morphemes by the Incels (as discussed shown in Table \ref{tab:morphemes}) also impacts performance, shown in the middle panel of Fig.~\ref{fig:results_task1}. These words are considered to be easier to identify based on their surface-level characteristics. Excluding the 570 positive terms that contain \textit{pill}, \textit{mog}, \textit{maxx}, \textit{cel}, \textit{chad}, \textit{cuck}, or \textit{oid}, we observe an average drop of 13.5\% in performance, which results in more similar scores in the best model for Incels (71.0\%) and Stormfront (64.4\%). 

\subsection{Task 2: Retrieval}
Results for the retrieval task are shown in Fig.~\ref{fig:results_task2} in App.~\ref{app:results}. We observe very similar trends compared to the classification formulation in Task 1: a consistent improvement in F1 with the full instructions, a drop in F1 when excluding the common subword cases, and higher scores for Incels compared to Stormfront. The absolute values are similar as well, with an average difference of 8.6\% and 7.4\% observed for the \texttt{10-examples} setting for Incels and Stormfront, respectively. This is a somewhat surprising result given the prior expectation that this is a more challenging task, although the same objective is evaluated in both. We observe that the labels coincide with the Task 1 results in 71\% (Incels) and 79\% (Stormfront) of cases. 

\subsection{Task 3: Decoding}\label{sec:llmm_def}
\begin{figure}[htb]
    \centering
    \includegraphics[width=\linewidth]{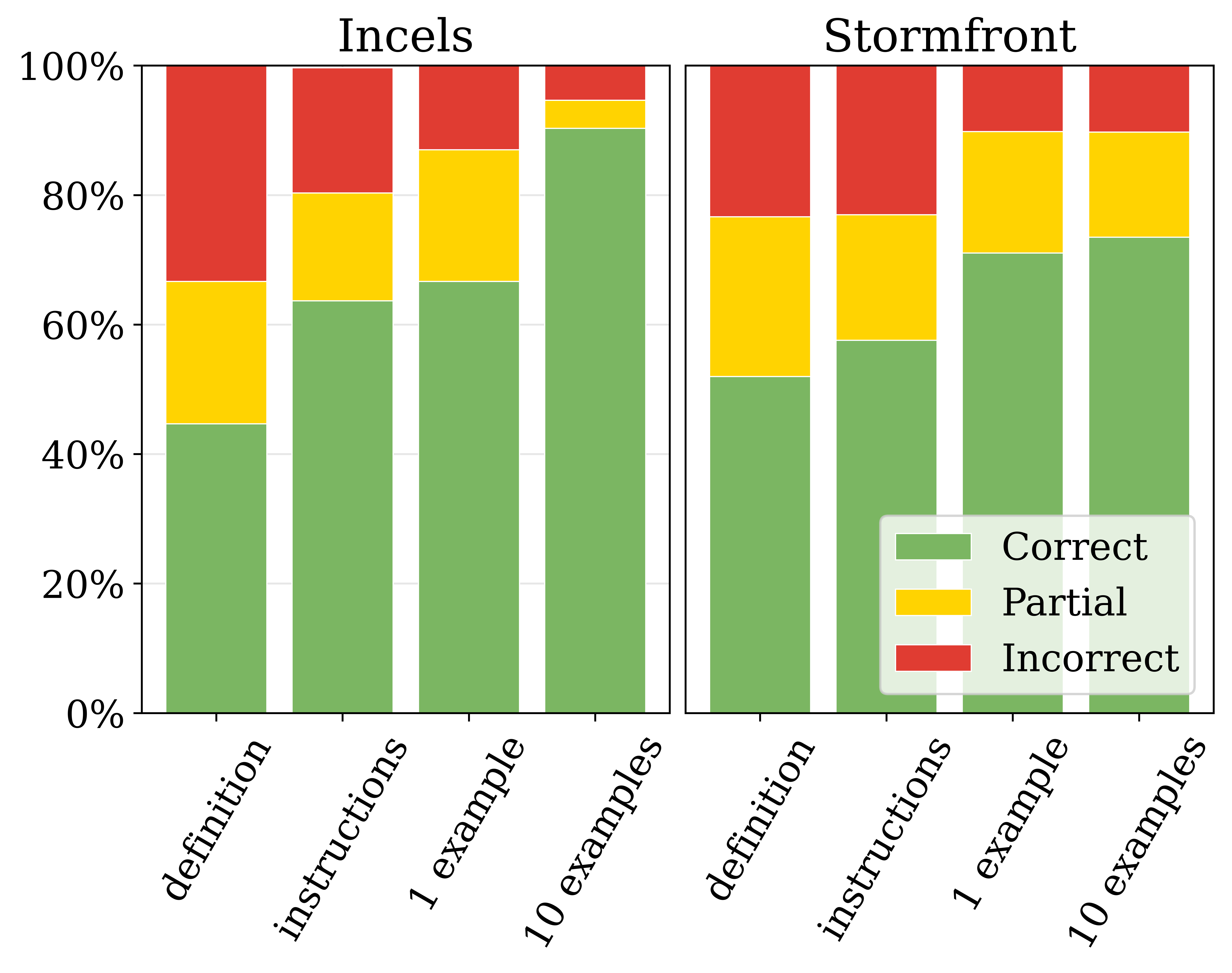}
    \caption{Results for Task 3 (Decoding).}
    \label{fig:task3}
\end{figure}

For the decoding task (Fig.~\ref{fig:task3}) we note a divergence from the results in Task 1 and 2, in that a substantial improvement is achieved over the \texttt{instructions} framing by including post examples. For Incels, a minor improvement is gained from \texttt{1-example} and a substantial increase from \texttt{10-examples}, generating correct definitions for 90.3\% of test cases, compared to 44.7\% in the \texttt{definition} framing. For Stormfront, a large improvement is observed when adding a single example post, but the \texttt{10-examples} framing does not provide a substantial further increase, generating correct definitions in 73.5\% of cases. We hypothesise that this is because the Stormfront language seems to be more heavily codified, often relying on acronyms or symbols. Considering the examples in Fig.~\ref{fig:posts_examples}, for example, the meaning of \textit{KLASP} could not be inferred from context alone as an acronym for \textit{Klannish Loyalty a Sacred Principle}\footnote{\url{https://www.adl.org/resources/hate-symbol/klasp}}. They also tend to use more symbolism; for instance, the use of triple parentheses is a well-known antisemitic dogwhistle \citep{mendelsohn2023dogwhistles}. By comparison, the Incels' language seems more focused on signalling in-group status rather than concealing its meaning from outsiders.
 
For both platforms, polysemous words are often incorrectly defined in lower-context framings. For example, \textit{fob} is an derogatory term for immigrants, but can also refer to a key fob. All prompts prime the model to produce the in-group sense, so these cases should still be viable. Similarly, providing only one post can result in the model ``overfitting'' to that use case, sometimes performing worse than without it. 
For the term \textit{apache}, a transphobic meme\footnote{\url{https://en.wikipedia.org/wiki/I_Sexually_Identify_as_an_Attack_Helicopter}}, all 10 examples refer to literal helicopters. Including more posts may mitigate this issue, but it is also an inherent challenge of this task. 
In a small number of cases, models respond with a variant of: \textit{I can't help with that request. The term provided contains hate speech and is not suitable for discussion}. This primarily occurs for the \texttt{definition} framing, which does not include the research motivation provided in the other framings. 

For this task, we do not observe a large difference in the results for the common subword cases for Incels (shown in Fig.~\ref{fig:task3_mrf} in App.~\ref{app:results}). This indicates that while the surface forms may be easier to identify, their meanings are not necessarily straightforward. For example, variants of \textit{-cel} are sometimes misinterpreted as referring to excelling. 

\section{Results for supervised classification}\label{sec:results_class}
Results for the supervised classification evaluation are shown in Table \ref{tab:macro_f1_combined_en}. We report F1 scores, averaged over 5 runs, for four different configurations for our two target tasks in English. In each case, the top row represents the baseline, i.e. the model of \citet{riabi-etal-2024-cloaked}. The columns represent different fine-tuning configurations, where each combining the target task with an auxiliary task. Statistical significance is determined using a one-sided $t$-test. 
The results demonstrate that domain adaptation on the IYKYK data significantly improves performance over the baseline model in most configurations for both tasks. We achieve state-of-the-art results on the {\tt Call for Action} task, improving over previously reported benchmarks. Interestingly, the combined forum data setting (denoted {\texttt{xlmt}\xspace}\textsubscript{+strf+inc}) frequently provides robust improvements beyond either of the platforms individually, despite the annotated data representing primarily Jihadist and alt-right ideologies. This suggests that merging data from diverse extremist sources enriches the models' capability to capture nuanced linguistic features common across radical communities.

\begin{table}[]
\footnotesize
\centering
\begin{subtable}{\linewidth}
\centering
\begin{tabular}{@{}p{1.05cm}p{1.25cm}|p{1.2cm}p{1.2cm}p{1.25cm}@{}}
\toprule
Model & CfA & +ideology & +NER & +rad.level \\
\midrule
 {\texttt{xlmt}\xspace} & 64.6\hphantom{$^\dagger$} & 65.1\hphantom{$^\dagger$} & 64.6\hphantom{$^\dagger$} & 64.8\hphantom{$^\dagger$} \\
 {\texttt{xlmt}\xspace}\textsubscript{+inc} & 68.1$^\dagger$\textcolor{green!60!black}{$_{(+3.5)}$} & 66.5\hphantom{$^\dagger$}\textcolor{green!60!black}{$_{(+1.4)}$} & 67.3$^\dagger$\textcolor{green!60!black}{$_{(+2.7)}$} & 67.4$^\dagger$\textcolor{green!60!black}{$_{(+2.6)}$} \\
{\texttt{xlmt}\xspace}\textsubscript{+strf} & 63.8\hphantom{$^\dagger$}\textcolor{red!60!black}{$_{(-0.9)}$} & 68.3$^\ddagger$\textcolor{green!60!black}{$_{(+3.2)}$} & 66.8$^\dagger$\textcolor{green!60!black}{$_{(+2.2)}$} & 67.7$^\dagger$\textcolor{green!60!black}{$_{(+2.9)}$} \\
{\texttt{xlmt}\xspace}\textsubscript{+strf+inc} & 66.7$^\dagger$\textcolor{green!60!black}{$_{(+2.0)}$} & 68.5$^\ddagger$\textcolor{green!60!black}{$_{(+3.4)}$} & 69.0$^\ddagger$\textcolor{green!60!black}{$_{(+4.4)}$} & 66.6\hphantom{$^\dagger$}\textcolor{green!60!black}{$_{(+1.7)}$} \\
\bottomrule
\end{tabular}
   \caption{Call for Action (CfA) prediction}\label{tab:cfa-results_en}
\end{subtable}
\begin{subtable}{\linewidth}
\centering
\begin{tabular}{@{}p{1.05cm}p{1.25cm}|p{1.2cm}p{1.2cm}p{1.25cm}@{}}
\toprule
 Model & Rad.level & +ideology & +NER & +CfA \\
\midrule
 {\texttt{xlmt}\xspace} & 56.4\hphantom{$^\dagger$} & 56.9\hphantom{$^\dagger$} & 58.3\hphantom{$^\dagger$} & 56.5\hphantom{$^\dagger$} \\
 {\texttt{xlmt}\xspace}\textsubscript{+inc} & 59.6\hphantom{$^\dagger$}\textcolor{green!60!black}{$_{(+3.2)}$} & 62.1$^\ddagger$\textcolor{green!60!black}{$_{(+5.2)}$} & 58.9\hphantom{$^\dagger$}\textcolor{green!60!black}{$_{(+0.6)}$} & 60.3$^\ddagger$\textcolor{green!60!black}{$_{(+3.8)}$} \\
 {\texttt{xlmt}\xspace}\textsubscript{+strf} & 59.6$^\dagger$\textcolor{green!60!black}{$_{(+3.2)}$} & 59.1$^\dagger$\textcolor{green!60!black}{$_{(+2.1)}$} & 59.7\hphantom{$^\dagger$}\textcolor{green!60!black}{$_{(+1.5)}$} & 60.6$^\ddagger$\textcolor{green!60!black}{$_{(+4.2)}$} \\{\texttt{xlmt}\xspace}\textsubscript{+strf+inc} & 60.0$^\dagger$\textcolor{green!60!black}{$_{(+3.6)}$} & 61.3$^\ddagger$\textcolor{green!60!black}{$_{(+4.3)}$} & 60.5\hphantom{$^\dagger$}\textcolor{green!60!black}{$_{(+2.2)}$} & 60.1$^\dagger$\textcolor{green!60!black}{$_{(+3.6)}$} \\

\bottomrule
\end{tabular}
\caption{Radicalization Level prediction}\label{tab:rad-results_en}
\end{subtable}
    \caption{Effect of domain adaptation on radical content classification in English. We use a one-sided t-test to determine statistical significance of improvements relative to the baseline, with $\dagger$ indicating $P$~<~0.05 and $\ddagger$ indicating $P$~<~0.01.}
\label{tab:macro_f1_combined_en}
\end{table} 
The combination of domain adaptation with auxiliary tasks through multi-task learning consistently achieves the best performance. Integrating ideology prediction alongside domain adaptation enhances the macro-F1 scores by up to 5.2\% for {\tt Radicalization Level} detection. 

We report the first results for the {\tt Radicalization Level} prediction task, establishing a baseline for future work. Predicting these labels is considered more challenging due to definitional ambiguity and contextual sensitivity, which is reflected in the comparatively lower performance scores observed in our experiments. 

Table~\ref{tab:cfa-results_fr} in App.~\ref{app:radical_content} shows the results for French. Compared to English, domain adaptation yields smaller gains. As the IYKYK data is primarily in English, the limited impact on French performance is expected.

\paragraph{Detecting hatespeech}
A central challenge for toxic language detection is the non-transferability of models across datasets due to labelling inconsistencies~\cite{fortunaHowWellHate2021,khondaker-etal-2023-cross} and vocabulary differences~\cite{rottgerHateCheckFunctionalTests2021}. Such concerns create particular challenges in applying toxicity and hatespeech detection technologies to communities which have high degrees of in-group language and have rapid linguistic evolution. In Table~\ref{tab:hate_results} in App.~\ref{app:results}, we test if hate speech detection models benefit from domain adaptation on the IYKYK data. We fine-tune the domain adapted models for classification on the dataset of \citet{montariol-etal-2022-multilingual}, using the same procedure as for radical content detection. Matching our expectations, performance on anti-immigrant hate speech increases when training on Stormfront, and performance on misogynist hate speech increases when training on Incels. This suggests that the insights in this paper may apply to toxic language detection problems more broadly.

\section{Discussion}
The results in Sec.~\ref{sec:tasks_results} and \ref{sec:results_class} indicate that it is necessary to first adapt NLP tools before application on extremist cryptolects. For in instruction-tuned models, this can be achieved by providing context details in the prompt. For encoder models, domain adaptation on data from extremist platforms significantly improves performance. These results have significant implications for automated content moderation systems, which often leverage LLMs\footnote{E.g., a \href{https://cloud.google.com/vertex-ai/generative-ai/docs/multimodal/gemini-for-filtering-and-moderation}{recommended usage} of the Gemini family of models in the official documentation.}. 

Toxic language is commonly removed during pre-training of NLP models to prevent the generation of such content~\cite{rae2022scalinglanguagemodelsmethods, anil2023palm2technicalreport, longpre-etal-2024-pretrainers, soldaini-etal-2024-dolma}. A consequence of such removals is that such tools are ill-suited for tasks relating to toxicity (e.g. modelling extremism) due to shifts in the underlying data distribution. Indeed, the removal of toxic content also has implications for the representation of other communities, such as queer communities in LLMs~\cite{xu-etal-2021-detoxifying}. 
Thus, two competing goals arise: on one hand, toxic language must not be generated by NLP models; on the other, the removal of toxic language during pre-training can limit the applicability of models to settings where good representations of toxicity and related concepts are paramount. 
This suggests a need for specialised models for generating content and for monitoring radical content and hate speech, respectively. The resources developed in this work provide a useful starting point for future work to develop such tools.

\section{Conclusion}
We conduct the first large-scale investigation of LLM performance on extremist cryptolects. Evaluating eight models over six tasks and two communities, our results provide useful insights for the use of LLMs in this domain. In particular, we find that the performance of instruction-tuned LLMs can be substantially improved by introducing additional context in the prompt. Furthermore, domain adaptation of encoder models on data from extremist platforms improves the detection of radical and toxic content. This study contributes to a limited but growing body of work in NLP on the topic of extremist language. We hope that the resources created in this work, including large datasets and expert-validated lexicons, will support further work in this important area. 

\section*{Ethics}
The ethical implications of research on radicalization require careful consideration. The research conducted in this work has been approved by an independent Ethics Review Board within our institution. It was considered to be a low-risk application since all data used in this work is publicly available and users interact on an anonymous basis. While we cannot guaranteed that users have not inadvertently provided personally identifiable information in their posts, the platforms that we have considered in this work discourage and block personal information where it is detected. In our discussion of the platforms' content in this paper, we do not include usernames or other personal details. 

We consider it important to ensure these datasets are accessible for research while avoiding unintended harms. While the models and datasets we develop can be used to support intervention and moderation to enhance social welfare, there is a risk of misuse, including the generation of toxic content and unjust profiling or preemptive prosecution. As such, we only make our models and datasets available to verified researchers upon request and under a memorandum of understanding requiring responsible use of the resources. Predictive models must be used carefully within broader intelligence systems, ensuring transparency and incorporating socio-cultural contexts to promote fair and responsible use \citep{Winter_Neumannt_2021}. 

Handling harmful text, such as hateful or radical content, poses challenges for researchers, annotators, and dataset curators. Repeated exposure to violent or hateful material can cause psychological harm, including vicarious trauma, highlighting the need for safeguards to protect those working with this data \citep{kirk-etal-2022-handling,riabi-etal-2024-cloaked}. In our evaluations, we only work with annotators for whom researcher risk management plans are in place. This includes consideration for mental health and personal safety aspects. In selecting examples for this paper, care was taken to ensure that the content is not too distressing to read, noting that there is a substantial psychological burden on researchers who work in this domain.

\section*{Limitations}
Since the trends we observe in our results are consistent over multiple communities, datasets and tasks, we believe our insights to be sound. However, we recognise that defining in-group language, like defining extremism, is subjective in nature. For this reason, we develop and share detailed annotation instructions, and perform pilot annotations to ensure alignment. We also prioritise high quality annotations by PhD-level experts. Even so, using a single annotator per dataset is a limitation; ideally, multiple annotations would be obtained for every sample. 

In this work, we do not account for the effect of time. We include posts from 2018 to 2024 in our analysis of in-group language. While our annotators have years of experience working on these groups, we recognise that the language in these groups is very dynamic and that the annotators may have a bias towards more recent language, which may play a role in their interpretations. 

Finally, many different extremist ideologies exist, and it is likely that the insights from this work would not fully apply to all of them. In fact, our results point to the fact that there are large differences in the performance for the two groups evaluated here. By including two communities in our study, we aim to provide a balanced impression and to focus on trends rather than on absolute values when interpreting our results. We hope that future work will use the framework proposed here to explore other ideologies.

\bibliography{custom}

\appendix

\section{Hyperparameters}\label{app:hyperparameters}

\subsection{Lexicon induction}\label{sec:lex_ind_hypers}
The LISTN-C method consists of \textit{(i)} jointly learning temporal user and word embeddings in a shared space, \textit{(ii)} clustering users into sub-groups to account for social fragmentation, and \textit{(iii)} extracting the maximum relevance words based on their embedding distance to the community centroid(s). We use the optimal configuration from \citet{de2024listn}, which assumes 5 clusters. While the original system investigates shorter-term linguistic change and therefore uses monthly data increments, we are interested in more stable terminology and therefore use annual snapshots. The original work studies lexical variation and therefore excludes standard words. Since we are also interested in dogwhistles, we maintain standard and non-standard words in our candidate set. We include only users active in more than one timestep and words used by more than 20 users. Following \citet{de2024listn}, we exclude non-alphabetic characters. Complete details of the system are provided in \citet{dekock2024jointlymodellingevolutioncommunity}.

\begin{table}[ht]
\centering
\footnotesize
\begin{tabular}{ll}
\toprule
\textbf{Hyperparameter} & \textbf{Value} \\
\midrule
K & 100 \\
$c_0$ & 0.005\\
$\lambda_1$ & 1 \\
$\lambda_2$ & 10 \\
Learning rate & 1e-05 \\
Batch size & 40 \\
Seed & 42 \\
Maximum epochs & 10000 \\
Optimizer & AdamW \\
Validation & 0.05 \\
Early stopping & P=100, tolerance=0.001\\
\bottomrule
\end{tabular}
\caption{Hyperparameters used for the \texttt{LISTN-C} system. Models are trained on a single NVIDIA GeForce GTX 108 GPU.}
\label{tab:mlm_hyperparameters}
\end{table}

\subsubsection{Domain Adaptation}
\begin{table}[ht]
\centering
\footnotesize
\begin{tabular}{ll}
\toprule
\textbf{Hyperparameter} & \textbf{Value} \\
\midrule
Learning rate & 8e-05 \\
Batch size & 32 \\
Seed & 42 \\
Distributed training type & Multi-GPU \\
Number of devices & 8 \\
Gradient accumulation steps & 8 \\
Total training batch size & 2048 \\
Number of epochs & 3 \\
Optimizer & AdamW \\
Learning rate scheduler & Linear \\
Warmup ratio & 0.01 \\
\bottomrule
\end{tabular}
\caption{Hyperparameters used for MLM training. We used the final checkpoint for all three models, selected based on performance on the development set. We conduct our experiments on RTX8000 GPUs.}
\label{tab:mlm_hyperparameters1}
\end{table}

\subsubsection{Supervised classification}

\begin{table}[htb]
\centering
\footnotesize
\begin{tabular}{ll}
\toprule
\textbf{Hyperparameter} & \textbf{Value} \\
\midrule
Encoder dropout & 0.2 \\
Update encoder weights & True \\
Batch size & 8 \\
Learning rate & 1e-4 \\
Weight decay & 0.01 \\
Optimizer betas & (0.9, 0.99) \\
Learning rate scheduler & Linear decay with warmup \\
Gradual unfreezing & True \\
Discriminative fine-tuning & True \\
Number of epochs & 10 \\
\bottomrule
\end{tabular}
\caption{Fine-tuning hyperparameters. The best checkpoint for each model is selected based on its performance on the development set. We conduct our experiments on RTX8000 GPUs.}
\label{tab:finetuning_hyperparams}
\end{table}

\section{Annotation instructions}\label{app:annotation_instructions}
\subsection{Identifying in-group language}\label{app:identification_instructions}
\subsubsection{Incels}
The goal of this work is to investigate in-group language within online extremist groups. In-group language is defined as specialised language developed by a social group to (1) indicate closeness to the in-group, and (2) create distance from the out-group. One way to think about it is: if you saw someone use this word in a neutral setting, e.g. Twitter, would it increase your expectation that they belong to an incel community? This would be strongly indicative, but is not a strict requirement for inclusion.

You will be given a list of words and a number of examples of these words being used on incels.is. These words have been identified as potential examples of in-group language to be used in a downstream analysis, which will compare the language in this community with a different group. It is expected that there will be invalid words in this list. We would like you to help us to filter them out. You don’t need to read all the examples; they are only provided as context.

Positive examples include:
\begin{itemize}\itemsep0em
\item Lexical innovations (non-standard words) invented by this group, e.g. foids.  Standard words can be thought of as what you might read in Wikipedia or a news source. 
\item Dogwhistles (standard words that have taken on a specific meaning to the group that is not held by the broader population), e.g. Becky, cope
\item Linguistic exports, or words that originated in the incel community but are now used more broadly, e.g. incel, mogging
\item Linguistic imports, or words that originated in another space and is elsewhere with the same meaning, but now belongs to the specialised language of this group, e.g., cuck, slut, hyoid, clavicles, gynocentric, alpha 
\item Names of people significant to the community
\end{itemize}

A word can belong to more than one of these categories.

Invalid words include:
\begin{itemize}\itemsep0em
\item  Standard language used in a standard way, e.g. internet
\item Internet language that is not specialised to this group, e.g. becuz
\item Spelling errors that are not specialised to this group, e.g. litterally.
\item Non-English text - there is a bit of this; please mark it as invalid.
\item Names of people or places, unless they are specifically relevant to this group. For example: Andrew Tate / Elliot Rodger = Yes, Anthony Albanese = Probably not.
\item Usernames (should be apparent from context).
\end{itemize}

This is a subjective task, and ultimately we are relying on your expert knowledge of this domain to make a judgment. Having said that, note that the idea is that we want to filter out words that are definitely invalid, so in the case of borderline words, you should opt to include them.

\subsubsection{Stormfront}
The goal of this work is to investigate in-group language within online extremist groups. In-group language is defined as specialised language developed by a social group to (1) indicate closeness to the in-group, and (2) create distance from the out-group. One way to think about it is: if you saw someone use this word in a neutral setting, e.g. Twitter, would it increase your expectation that they belong to an alt-right community? This would be strongly indicative, but is not a strict requirement for inclusion.

You will be given a list of words and a number of examples of these words being used on Stormfront.org. These words have been identified as potential examples of in-group language to be used in a downstream analysis, which will compare the language in this community with a different group. It is expected that there will be invalid words in this list. We would like you to help us to filter them out. You don’t need to read all the examples; they are only provided as context.

Positive examples include:
\begin{itemize}\itemsep0em
\item Lexical innovations (non-standard words) invented by this group, e.g. foids.  Standard words can be thought of as what you might read in Wikipedia or a news source. 
\item Dogwhistles (standard words that have taken on a specific meaning to the group that is not held by the broader population), e.g. Becky, cope
\item Linguistic exports, or words that originated in the alt-right community but are now used more broadly, e.g. incel, mogging
\item Linguistic imports, or words that originated in another space and is elsewhere with the same meaning, but now belongs to the specialised language of this group, e.g., cuck, slut, hyoid, clavicles, gynocentric, alpha 
\item Names of people significant to the community
\end{itemize}

A word can belong to more than one of these categories.

Invalid words include:
\begin{itemize}\itemsep0em
    \item Standard language used in a standard way, e.g. internet
    \item  Internet language that is not specialised to this group, e.g. becuz
    \item  Spelling errors that are not specialised to this group, e.g. litterally.
    \item  Non-English text - there is a bit of this; please mark it as invalid.
    \item  Names of people or places, unless they are specifically relevant to this group. For example: Andrew Tate / Elliot Rodger = Yes, Anthony Albanese = Probably not.
    \item  Usernames (should be apparent from context).
\end{itemize}

This is a subjective task, and ultimately we are relying on your expert knowledge of this domain to make a judgment. Having said that, note that the idea is that we want to filter out words that are definitely invalid, so in the case of borderline words, you should opt to include them.

\subsection{Defining in-group language}\label{app:define_instructions}
\subsubsection{Incels}
Given a word and a set of 3 definitions, assess the validity of each definition in the context of Incels.is. 

These definitions are generated by a large language model. The goal is to investigate the quality of the definitions generated using different approaches. The definitions for each word are randomly ordered and does not correspond to the same approach every time. 

NOTE: More than one definition can be correct (or incorrect, or partially correct etc) for each word. There does not need to be a “best” one.

Possible answers are:
\begin{itemize}\itemsep0em
\item Accurate and complete: a valid definition that captures all important aspects of the term.
\item Partially correct: Correct in some ways but missing some nuance. Note: it is preferable not to use this option if one of the others might apply. 
\item Inaccurate: Does not capture the meaning of the word in this context. 
\item No Answer: Model refuses to answer or provides a nonsensical output, e.g. ""I cannot help you with that."" or ""<your definition>""		
\end{itemize}

\subsubsection{Stormfront}
Given a word and a set of 3 definitions, assess the validity of each definition in the context of Stormfront.org. 

These definitions are generated by a large language model. The goal is to investigate the quality of the definitions generated using different approaches. The definitions for each word are randomly ordered and does not correspond to the same approach every time. 

NOTE: More than one definition can be correct (or incorrect, or partially correct etc) for each word. There does not need to be a “best” one.

Possible answers are:
\begin{itemize}\itemsep0em
\item Accurate and complete: a valid definition that captures all important aspects of the term.
\item Partially correct: Correct in some ways but missing some nuance. Note: it is preferable not to use this option if one of the others might apply. 
\item Inaccurate: Does not capture the meaning of the word in this context. 
\item No Answer: Model refuses to answer or provides a nonsensical output, e.g. ``I cannot help you with that.'' or ``<your definition>''		
\end{itemize}

\section{Prompts}\label{app:prompts}
For each model, we use the appropriate prompt format as per the model card and a temperature of 0.3. Tests are run through \url{together.ai} for a total of \$154 in API costs. Outputs are parsed by extracting the text following \textit{ANSWER}. In addition to the instructions below, we also provide the required output format and demarcations of the instructions and task.

\subsection{Task 1: Classification}
\subsubsection{Definition}
\textit{In-group language is defined as specialised language developed by a social group to  (1) indicate closeness to the in-group, and (2) create distance from the out-group. You will be given a word. Your job is to determine if it is in-group language.}
\subsubsection{Instructions, 1-example and 10-examples}
The prompt is constructed based on the instructions in Sec.~\ref{app:define_instructions}, preceded by: \textit{You are an expert sociologist working on online extremist groups}. Then, we provide either no examples (referred to as \textit{instructions}), a single post (\textit{1-example}) or a list of ten (newline-separated) posts (\textit{10-examples}).

\subsection{Task 2: Retrieval}
\subsubsection{Definition}
\textit{In-group language is defined as specialised language developed by a social group to (1) indicate closeness to the in-group, and (2) create distance from the out-group. You will be given a post from an online forum. Your task is to identify instances of in-group language in this post. Return a list of words that are positive examples according to the description of in-group language provided. If there are none, return an empty list.}

\subsubsection{Instructions}
The prompt for this task consists of the instructions in Sec.~\ref{app:annotation_instructions}, preceded by: \textit{You are an expert sociologist working on online extremist groups}. This instructions are followed by: \textit{You will be given a post from [Stormfront.org | Incels.is]. Your task is to identify instances of in-group language in this post. Return a list of words that are positive examples according to the description of in-group language provided. If there are none, return an empty list.}

\subsection{Task 3: Decoding}
\subsubsection{Definition} \textit{The goal of this work is to investigate in-group language within online groups. You will be given a word used in an online conversation. This word has been identified as an example of in-group language, which includes non-standard language. Your task is to infer the meaning of the word. }

\subsubsection{Instructions, 1-example, and 10-examples}
The prompt is constructed based on the following description, followed by one, ten or zero examples: \textit{The goal of this work is to investigate in-group language within online extremist groups. You will be given a word used on [Stormfront.org, an online alt-right community | Incels.is, an online anti-women community]. This word has been identified as an example of in-group language, which includes non-standard language. 
Your task is to infer the meaning of the word.}

\section{Additional results}\label{app:results}

\subsection{Zero-shot evaluation}
\begin{figure*}[htb]
    \centering
    \includegraphics[width=\linewidth]{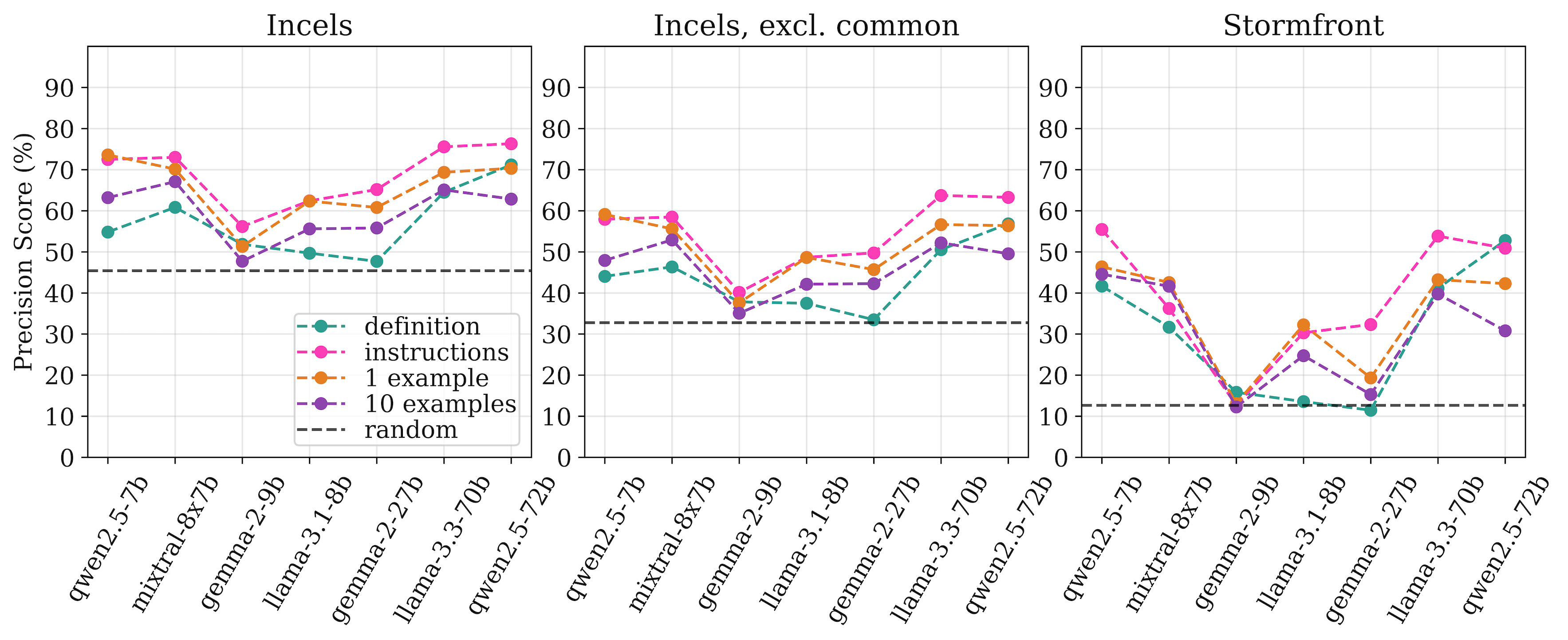}
    \caption{Precision scores for Task 1.}
    \label{fig:task1_precision}
\end{figure*}

\begin{figure*}[htb]
    \centering
    \includegraphics[width=\linewidth]{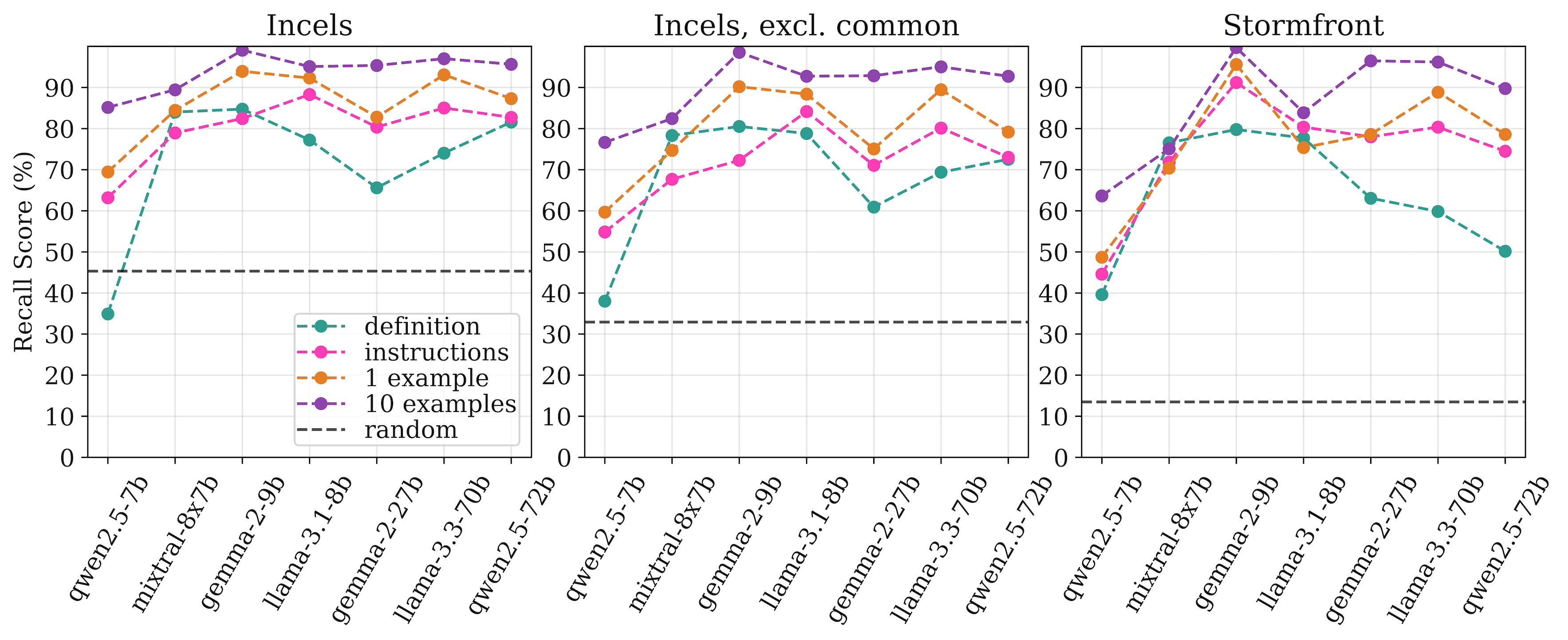}
    \caption{Recall scores for Task 1.}
    \label{fig:task1_recall}
\end{figure*}

\begin{figure*}[t]
    \centering
    \includegraphics[width=\linewidth]{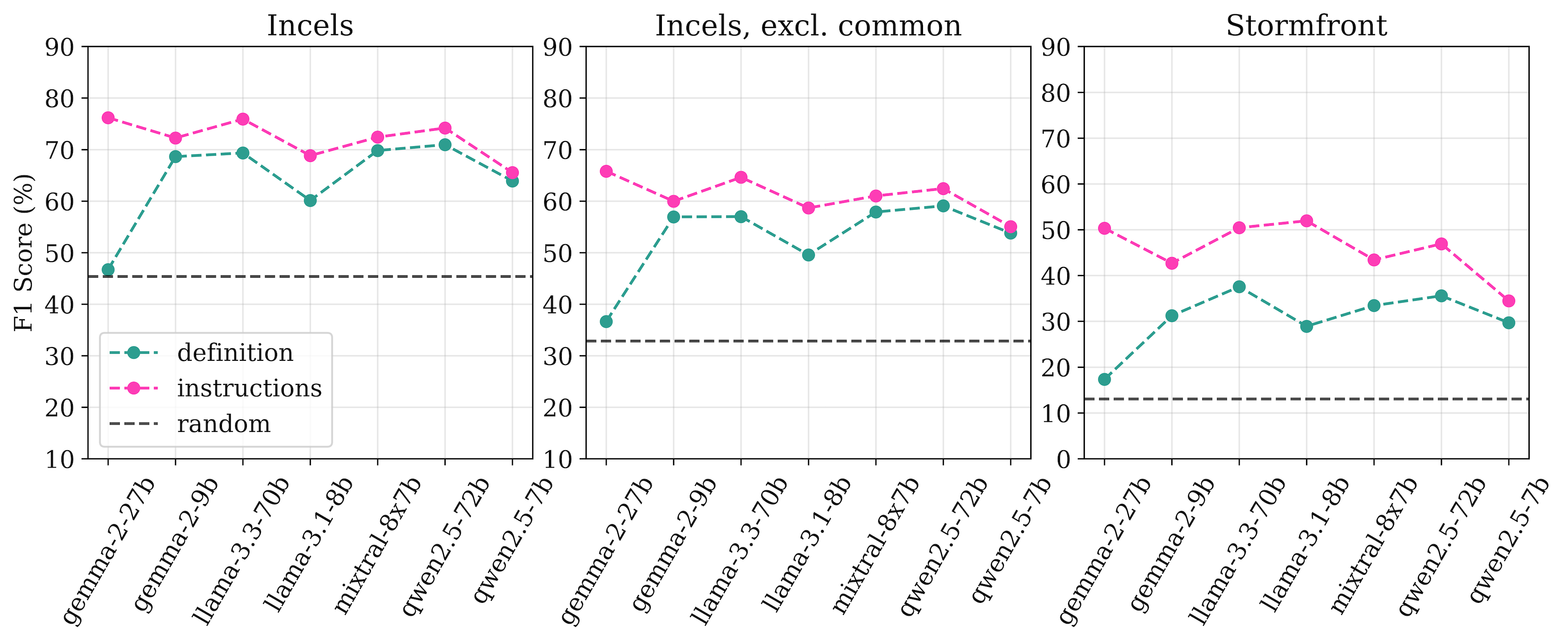}
    \caption{Results for Task 2 (Retrieval).}
    \label{fig:results_task2}
\end{figure*}

\begin{figure}[htb]
    \centering
    \includegraphics[width=\linewidth]{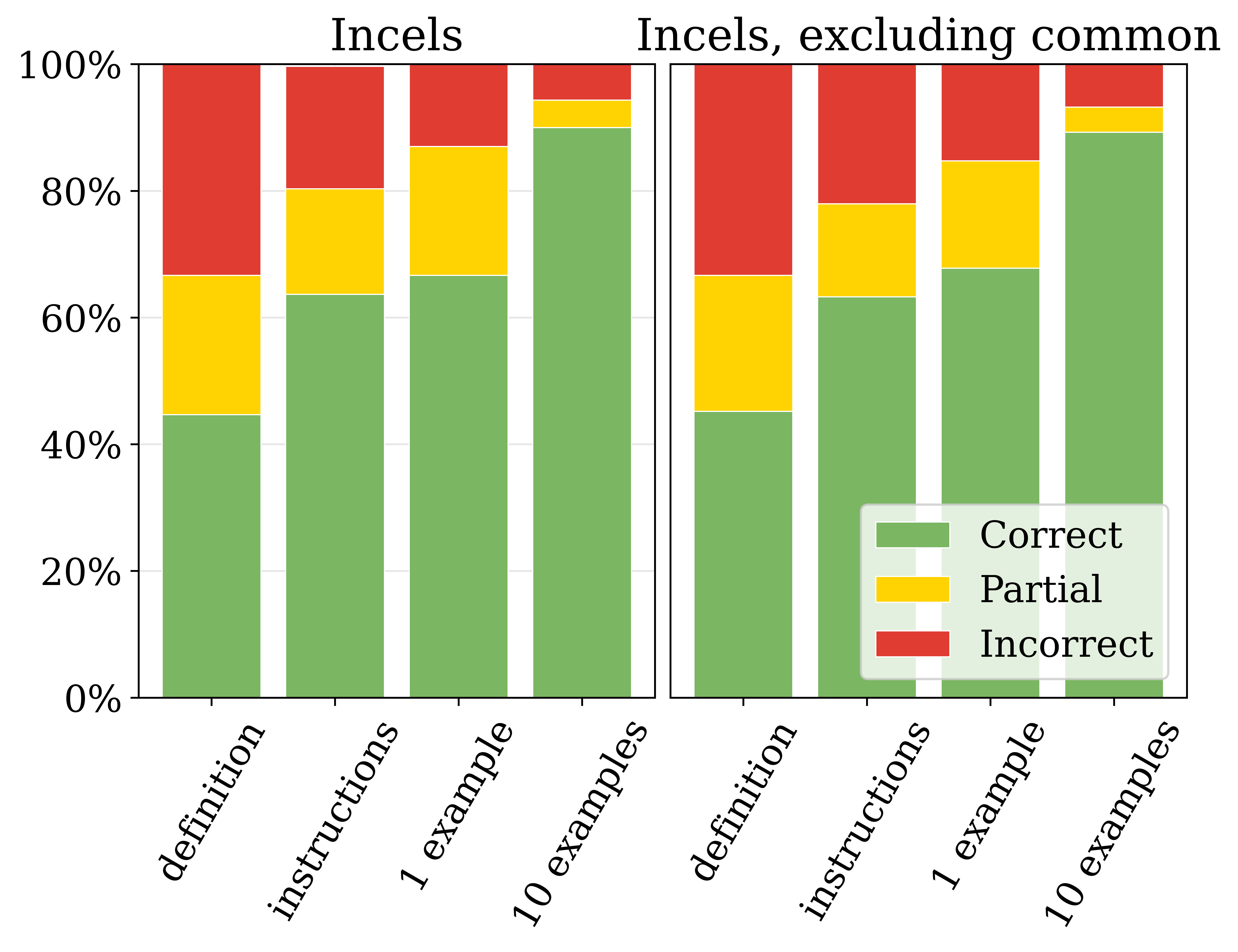}
    \caption{Results for Task 3, excluding common morphemes for Incels.}
    \label{fig:task3_mrf}
\end{figure}

Precision and recall scores for Task 1 are shown in Figures \ref{fig:task1_precision} and \ref{fig:task1_recall}, respectively. Fig.~\ref{fig:results_task2} contains the F1 scores for Task 2. The results for the common subwords variant for Incels for Task 3 are shown in Fig.~\ref{fig:task3_mrf}.

\subsection{Detecting radical content}
Results for the radical content detection task in French are provided in Tables \ref{tab:cfa-results_fr} and \ref{tab:macro_f1_combined_fr}. Results for the hatespeech detection task are in Table \ref{tab:hate_results}.

\label{app:radical_content}
\begin{table*}[htb!]
\footnotesize
\centering
\begin{subtable}{1\linewidth}
\centering
\begin{tabular}{lcccc}
\toprule
 Model & CfA & CfA+ideology & CfA+NER & CfA+rad. level \\
\midrule
 xlmt & 65.7\hphantom{$^\dagger$} & 65.6\hphantom{$^\dagger$} & 62.0\hphantom{$^\dagger$} & 63.9\hphantom{$^\dagger$} \\
xlmt\textsubscript{+incels} & 65.0\hphantom{$^\dagger$}\textcolor{red!60!black}{$_{(-0.6)}$} & 68.7\hphantom{$^\dagger$}\textcolor{green!60!black}{$_{(+3.2)}$} & 63.8\hphantom{$^\dagger$}\textcolor{green!60!black}{$_{(+1.9)}$} & 64.2\hphantom{$^\dagger$}\textcolor{green!60!black}{$_{(+0.3)}$} \\
 xlmt\textsubscript{+stormfront}  & 65.9\hphantom{$^\dagger$}\textcolor{green!60!black}{$_{(+0.3)}$} & 62.3\hphantom{$^\dagger$}\textcolor{red!60!black}{$_{(-3.3)}$} & 61.7\hphantom{$^\dagger$}\textcolor{red!60!black}{$_{(-0.3)}$} & 62.5\hphantom{$^\dagger$}\textcolor{red!60!black}{$_{(-1.4)}$} \\
 xlmt\textsubscript{+stormfront+incels} & 65.8\hphantom{$^\dagger$}\textcolor{green!60!black}{$_{(+0.2)}$} & 64.0\hphantom{$^\dagger$}\textcolor{red!60!black}{$_{(-1.6)}$} & 63.8\hphantom{$^\dagger$}\textcolor{green!60!black}{$_{(+1.8)}$} & 59.4\hphantom{$^\dagger$}\textcolor{red!60!black}{$_{(-4.5)}$} \\

\bottomrule
\end{tabular}
   \caption{Call for Action (CfA) prediction}\label{tab:cfa-results_fr}
\end{subtable}
\bigskip
\begin{subtable}{1\linewidth}
\centering
\begin{tabular}{lcccc}
\toprule
Model & Rad. level & Rad. level+ideology & Rad. level+NER & Rad. level+CfA \\
\midrule
 xlmt & 61.7\hphantom{$^\dagger$} & 59.8\hphantom{$^\dagger$} & 56.9\hphantom{$^\dagger$} & 59.1\hphantom{$^\dagger$} \\
 xlmt\textsubscript{+incels} & 60.0\hphantom{$^\dagger$}\textcolor{red!60!black}{$_{(-1.7)}$} & 60.0\hphantom{$^\dagger$}\textcolor{green!60!black}{$_{(+0.2)}$} & 57.7\hphantom{$^\dagger$}\textcolor{green!60!black}{$_{(+0.8)}$} & 58.4\hphantom{$^\dagger$}\textcolor{red!60!black}{$_{(-0.7)}$} \\
 xlmt\textsubscript{+stormfront} & 57.8\hphantom{$^\dagger$}\textcolor{red!60!black}{$_{(-3.9)}$} & 60.2\hphantom{$^\dagger$}\textcolor{green!60!black}{$_{(+0.4)}$} & 57.3\hphantom{$^\dagger$}\textcolor{green!60!black}{$_{(+0.4)}$} & 58.1\hphantom{$^\dagger$}\textcolor{red!60!black}{$_{(-1.0)}$} \\
 xlmt\textsubscript{+stormfront+incels} & 58.8\hphantom{$^\dagger$}\textcolor{red!60!black}{$_{(-2.9)}$} & 58.4\hphantom{$^\dagger$}\textcolor{red!60!black}{$_{(-1.3)}$} & 57.6\hphantom{$^\dagger$}\textcolor{green!60!black}{$_{(+0.8)}$} & 58.2\hphantom{$^\dagger$}\textcolor{red!60!black}{$_{(-0.9)}$} \\
\bottomrule
\end{tabular}
\caption{Radicalization Level prediction}\label{tab:rad-results_fr}
\end{subtable}

    \caption{Effect of domain adaptation on radical content classification macro-F1 scores (\%) for French, averaged over 5 runs. 
Columns represent different fine-tuning configurations, where each setting combines the target task 
(Call for Action or Radicalization Level detection) with an auxiliary task to assess its influence on performance. 
Main scores are shown in black, with the subscript indicating the delta relative to the \texttt{xlmt} baseline. 
\textcolor{green!60!black}{Green} subscripts denote an improvement over the baseline, 
\textcolor{red!60!black}{red} subscripts a drop. Statistical significance is determined using a one-sided $t$-test 
on the five run scores. A dagger ($\dagger$) indicates $p$~<~0.05, and a double-dagger ($\ddagger$) indicates $p$~<~0.01.
}
\label{tab:macro_f1_combined_fr}
\end{table*}

\begin{table}[]
    \centering
\begin{tabular}{llcc}
\toprule
 Model & immigrants & women \\
\midrule
 xlmt & 77.0\hphantom{$^\dagger$} & 78.8\hphantom{$^\dagger$} \\
 xlmt\textsubscript{+incels} & 76.0\hphantom{$^\dagger$}\textcolor{red!60!black}{$_{(-0.9)}$} & 79.9\hphantom{$^\dagger$}\textcolor{green!60!black}{$_{(+1.1)}$} \\
xlmt\textsubscript{+stormfront} & 78.1\hphantom{$^\dagger$}\textcolor{green!60!black}{$_{(+1.1)}$} & 78.8\hphantom{$^\dagger$}\textcolor{red!60!black}{$_{(-0.0)}$} \\
 xlmt\textsubscript{+stormfront+incels} & 76.9\hphantom{$^\dagger$}\textcolor{red!60!black}{$_{(-0.0)}$} & 79.4\hphantom{$^\dagger$}\textcolor{green!60!black}{$_{(+0.6)}$} \\
\bottomrule
\end{tabular}
    \caption{Hate speech detection macro-F1 scores on the English using corpora from HatEval and the splitting from \citet{montariol-etal-2022-multilingual}. We use same experimental setting for supervised classification as radicalization detection.}
    \label{tab:hate_results}
\end{table}
\end{document}

%% file: examples.tex
\usetikzlibrary{positioning, shapes, fit, calc}
\tikzset{
    textbox/.style={
        draw,
        %dashed,
        rounded corners,
        inner sep=3pt,
        text width=0.85\linewidth,
        font=\normalsize,
        align=left
    },
    greybox/.style={
        textbox,
        fill=gray!10,
        draw=gray!0
    },
    whitebox/.style={
        textbox,
        %dashed,
        fill=green!10,
        draw=gray!0
    }
}

\begin{tikzpicture}
        \node[greybox] (box1) {
            \textbf{1. }the movie is about \textcolor{magenta}{normies} \textcolor{blue}{roping} and \textcolor{orange}{mentalcels} \textcolor{teal}{going ER}
        };
        \node[whitebox, below=0.2cm of box1, xshift=0.7cm] (box1B) {
            The movie is about \textcolor{magenta}{average people [\ding{51}]} \textcolor{blue}{forming romantic relationships [\ding{55}]} and \textcolor{orange}{intellectually-inclined, socially isolated individuals [\ding{55}]} \textcolor{teal}{experiencing extreme emotional distress or rage [\ding{55}]}.
        };
        \draw[->, thick, purple!70, dashed] ($(box1.south west)+(0.25cm,0)$) -- ++(0,-1cm) -- ++(0.45cm,0cm);

        \node[text=purple!70, font=\Large] at ($(box1.south west)+(0.25,-0.5cm)$) {\faRobot};

        \node[greybox, below=2.85cm of box1] (box2) {
            \textbf{2.} You should \textcolor{blue}{gymcel} to \textcolor{magenta}{looksmaxx} and \textcolor{orange}{healthmaxx} primarily.
        };
        \node[whitebox, below=0.2cm of box2, xshift=0.7cm] (box2B) {
            You should focus on \textcolor{magenta}{improving your physical appearance [\ding{51}]} and overall \textcolor{orange}{health [\ding{51}]} through \textcolor{blue}{exercise [\ding{51}]}.
        };
        \draw[->, thick, purple!70, dashed] ($(box2.south west)+(0.25cm,0)$) -- ++(0,-1cm) -- ++(0.45cm,0cm);

        \node[text=purple!70, font=\Large] at ($(box2.south west)+(0.25,-0.5cm)$) {\faRobot};

        \node[greybox, below=2cm of box2] (box3) {
            \textbf{3.} \textcolor{orange}{Kigy} im looking for a \textcolor{magenta}{Klan} that I can call home. \textcolor{blue}{KLASP}
        };
        \node[whitebox, below=0.2cm of box3, xshift=0.7cm] (box3B) {
            \textcolor{orange}{Klansman I Greet You [\ding{51}]}, I am looking for a \textcolor{magenta}{Ku Klux Klan group [\ding{51}]} that I can join and feel a sense of belonging. \textcolor{blue}{Keep Loyalty And Safety Paramount [\ding{55}]}
        };
        \draw[->, thick, purple!70, dashed] ($(box3.south west)+(0.25cm,0)$) -- ++(0,-1cm) -- ++(0.45cm,0cm);

        \node[text=purple!70, font=\Large] at ($(box3.south west)+(0.25,-0.5cm)$) {\faRobot};

    \end{tikzpicture}